\documentclass[conference]{IEEEtran}
\IEEEoverridecommandlockouts

\usepackage{cite}
\usepackage{graphics} 
\usepackage{epsfig} 
\usepackage{mathptmx} 
\usepackage{times} 
\usepackage{amsmath} 
\usepackage{amssymb}  
\usepackage{xcolor}
\usepackage{color}
\usepackage[linesnumbered,ruled,vlined]{algorithm2e}
\usepackage{algorithmic}
\usepackage{subfigure}
\usepackage{multirow}
\usepackage{float}
\usepackage{algorithm2e}
\usepackage{booktabs}
\usepackage[utf8x]{inputenc}
\usepackage{wrapfig}
\usepackage{threeparttable}
\usepackage{boldline}
\usepackage{url}
\usepackage{bm}

\usepackage{xcolor}

\renewcommand{\Re}{\mathbb{R}}

\DeclareMathAlphabet{\mathcal}{OMS}{cmsy}{m}{n}
\renewcommand{\baselinestretch}{1}

\SetKwComment{Comment}{$\triangleright$\ }{}

\title{\LARGE \bf
    Obstacle- and Occlusion-Responsive Visual Tracking Control for \\
    Redundant Manipulators using Reachability Measure
}

\author{Mincheul Kang$^{1}$ and Junhyoung Ha$^{2}$, \IEEEmembership{Member,~IEEE}
    \thanks{This work was supported by the Korea Institute of Science and Technology (KIST) Institutional Program under the MIDAS2 grant (2E32272).}
    \thanks{{$^1$}Mincheul Kang is with Mechatronics Research and Development Center, Samsung Electronics, Hwaseong-si 18448, South Korea (e-mail: {\tt\small kmc050210@gmail.com}).}
    \thanks{{$^2$}Junhyoung Ha is with the Center for Healthcare Robotics, Artificial Intelligence and Robotics Institute, Korea Institute of Science and Technology, Seoul 02792, South Korea (e-mail: {\tt\small hjhdog1@gmail.com}).}
    \thanks{This work has been submitted to the IEEE for possible publication. Copyright may be transferred without notice, after which this version may no longer be accessible.}
}

\newcommand{\Skip}[1]{}
\renewcommand{\paragraph}[1]{{\bf {#1}}}  

\def\HiLi{\leavevmode\rlap{\hbox to 
        \hsize{\color{yellow!50}\leaders\hrule height .8\baselineskip depth .5ex\hfill}}}

\begin{document}
    \maketitle
    \thispagestyle{empty}
    \pagestyle{empty}
    
    \newtheorem{thm}{Theorem}
    \newtheorem{lem}[thm]{Lemma}
    \newtheorem{col}[thm]{Corollary}


\begin{abstract}
A vision system attached to a manipulator excels at tracing a moving target object while effectively handling obstacles, overcoming limitations arising from the camera's confined field of view and occluded line of sight. Meanwhile, the manipulator may encounter certain challenges, including restricted motion due to kinematic constraints and the risk of colliding with external obstacles. These challenges are typically addressed by assigning multiple task objectives to the manipulator. However, doing so can cause an increased risk of driving the manipulator to its kinematic limits, leading to failures in object tracking or obstacle avoidance. To address this issue, we propose a novel visual tracking control method for a redundant manipulator that takes the kinematic constraints into account via a reachability measure. Our method employs an optimization-based controller that considers object tracking, occlusion avoidance, collision avoidance, and the kinematic constraints represented by the reachability measure.
Subsequently, it determines a suitable joint configuration through real-time inverse kinematics, accounting for dynamic obstacle avoidance and the continuity of joint configurations. To validate our approach, we conducted simulations and hardware experiments involving a moving target and dynamic obstacles. The results of our evaluations highlight the significance of incorporating the reachability measure.
\end{abstract}


\section{Introduction}
\label{sec:1}

Visual Tracking Systems (VTSs) have been utilized in various fields, including robotics, virtual reality, surgical navigation, and person identification~\cite{van2003optical, han2021fast, cho2022part, saeedi2023automatic}. In robot-assisted surgical navigation, VTSs are actively used to track the location of surgical instruments~\cite{kim2017registration, herregodts2021improved, xu2022information}. A significant challenge in these applications is effectively dealing with obstacles that impede object tracking.

Cameras are vulnerable to occlusion caused by obstacles due to line of sight (LoS) constraints~\cite{xu2022information}. Additionally, cameras can only track objects within their field of view (FoV). To address these limitations, two methods have been proposed: (i) multi-camera integration and (ii) active camera navigation using a robotic platform.

Garc{\'\i}a-V{\'a}zquez et al.~\cite{garcia2013feasibility} proposed a multi-camera visual tracking system to address the occlusion problem caused by obstacles. Wang et al.~\cite{wang2015towards} developed a reconfigurable monocular system with a fast calibration method to make cameras easily movable. Also, Dai et al.~\cite{dai2020prior} improved the calibration accuracy of multi-camera systems utilizing a neural network with learned prior knowledge. While these multi-camera methods can alleviate occlusion problems, they require significant computational power and additional space for system construction.

Kuo et al.~\cite{kuo2005development} proposed an active camera navigation system that involves mounting a camera on a manipulator. The manipulator is then controlled to dynamically track a moving object. However, this method did not account for occlusion caused by obstacles. Wang et al.~\cite{wang2015robot} recognized this limitation and proposed an occlusion avoidance method using a four-degree-of-freedom (DoF) robotic platform installed on the ceiling. Meng et al.~\cite{meng2021development} also tackled the occlusion problem by employing a multi-DoF manipulator. These methods excel in tracking objects even when they are occluded by obstacles or lie outside the camera's FoV. Nevertheless, the inherent risk of collision with obstacles when using a manipulator has not been thoroughly explored.

\begin{figure}[t]
	\vspace{0.2cm}
	\centering
	\includegraphics[width=3.25in]{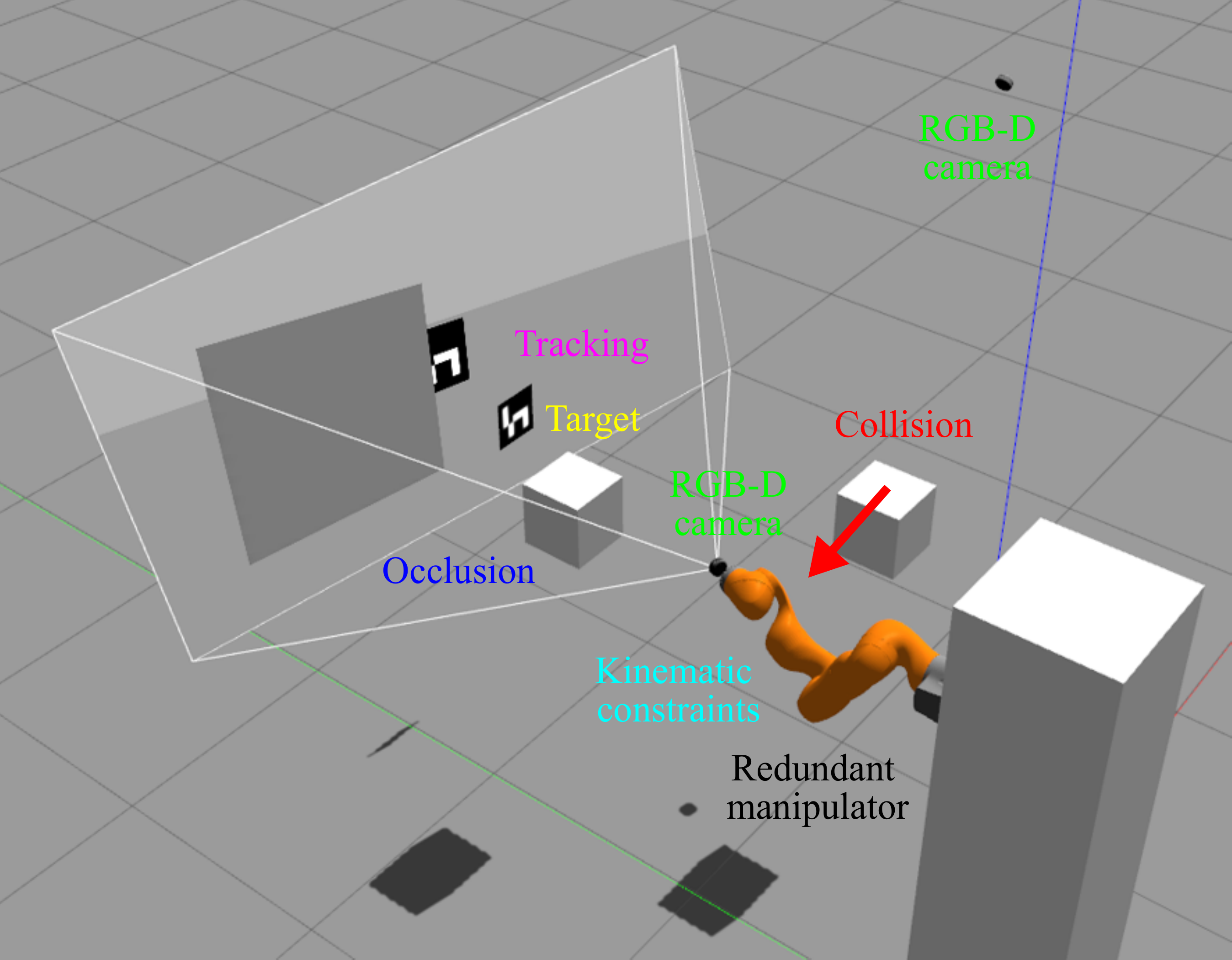}
	\caption{
            Problem definition: a redundant manipulator combined with a camera is tracking a moving target while avoiding occlusions and obstacles. An external RGB-D camera is utilized to help identify obstacles. During manipulator control, we take into account various factors, including object tracking, avoiding occlusions and collisions, and adhering to the manipulator's kinematic constraints such as joint position limits and restricted workspace induced by static environment.
	}
	\label{fig:main}
	\vspace{-0.6cm}
\end{figure}

\begin{figure*}[t]
	\vspace{0.3cm}
	\centering
	\subfigure [Limited FoV] {
		\includegraphics[width=1.62in]{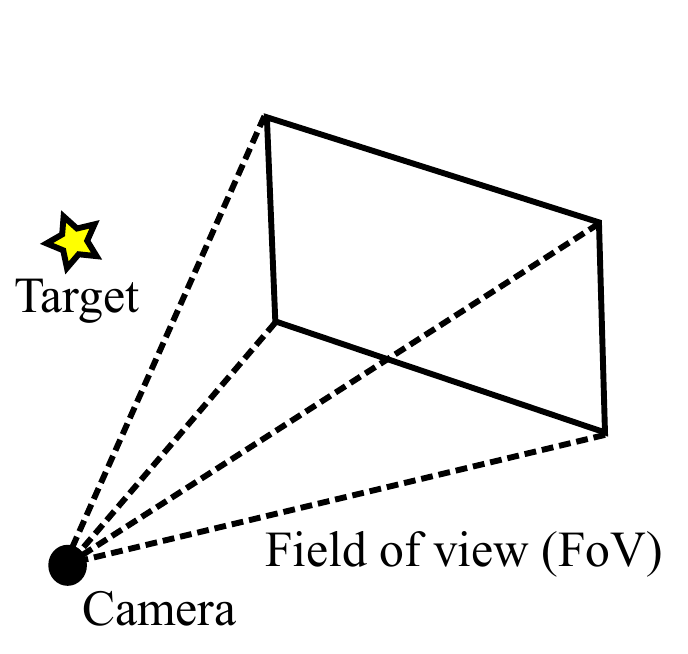}
		\label{fig:fov} 
	} 
	\subfigure [Occlusion] { 
		\includegraphics[width=1.62in]{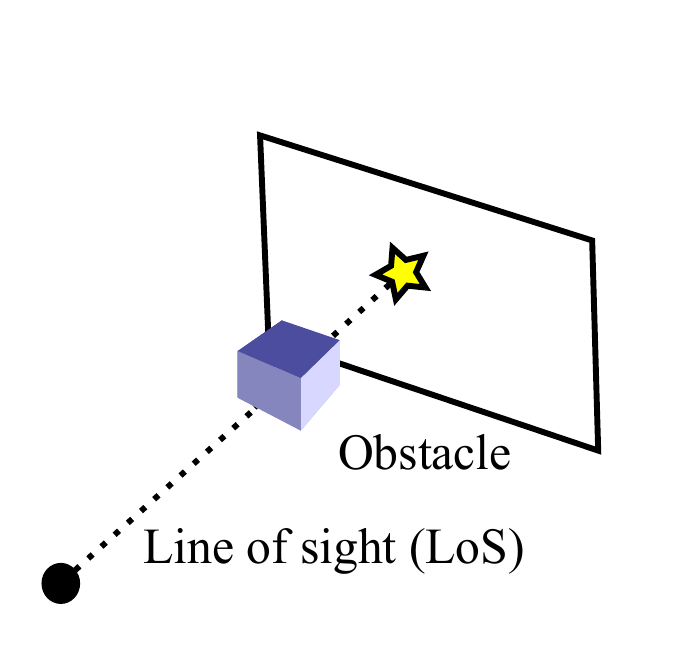}
		\label{fig:occlusion} 
	}
        \subfigure [Collision] {
		\includegraphics[width=1.62in]{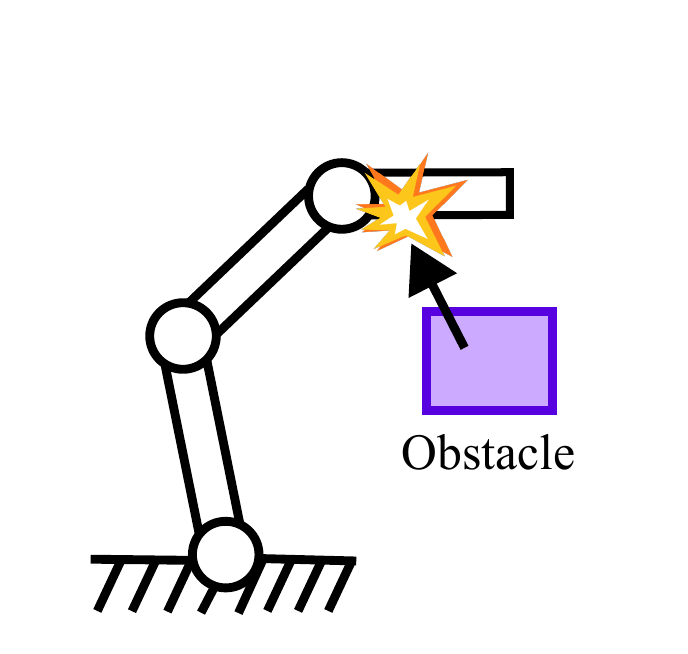}
		\label{fig:collision} 
	} 
	\subfigure [Kinematic constraint] { 
		\includegraphics[width=1.62in]{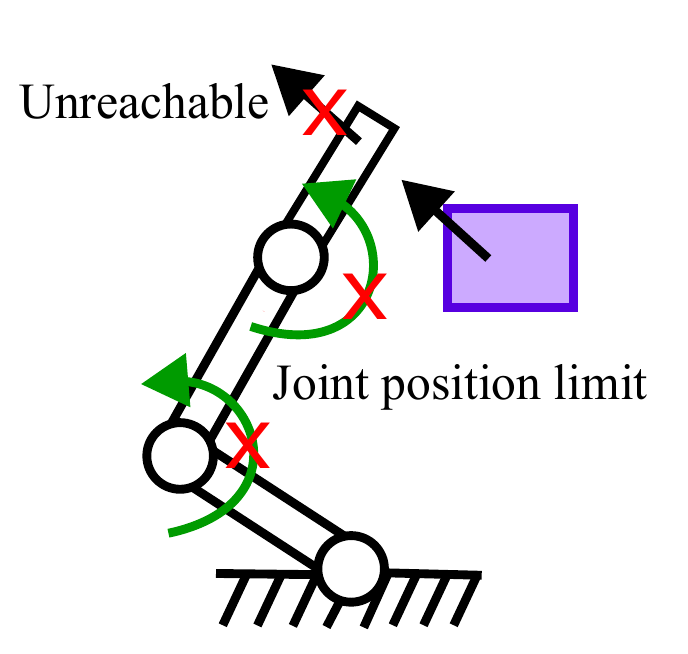}
		\label{fig:kinematic_constraint} 
	}
	\caption{
		  Illustrations of four challenges in our visual tracking problem.
	}
	\label{fig:problems}
	\vspace{-0.4cm}
\end{figure*}

Several manipulation methods have been proposed for real-time collision avoidance of dynamic obstacles in the context of manipulator control~\cite{hauser2012responsiveness, rakita2021collisionik, kang2021rcik}. However, the integration of visual tracking tasks with obstacle avoidance has not been fully explored. More precisely, during the process of avoiding collision and occlusion by dynamic obstacles and tracking a moving object simultaneously, the manipulator is prone to reaching its kinematic limits, such as joint position limits and restricted workspace. These constraints can lead to potential tracking failure or collisions with obstacles.

To address this research gap, in this letter, we present an occlusion- and obstacle-responsive visual tracking controller for a manipulator-driven visual tracking system, which takes into account the manipulator's kinematic characteristics to intelligently navigate around collisions and occlusions. We utilize a redundant manipulator for its flexibility in motion. Our method can be outlined as follows:
\begin{itemize}
    \item We employ a redundant manipulator equipped with an RGB-D camera mounted on its end-effector. Additionally, an external RGB-D camera is used to observe the manipulator to identify surrounding obstacles. See Fig.~\ref{fig:main} for an illustration.
    \item A non-linear optimization approach is utilized to determine the desired end-effector pose, considering four distinct objectives: (i) object tracking, (ii) occlusion avoidance, (iii) collision avoidance, and (iv) kinematic constraints (represented by a reachability measure~\cite{zacharias2007capturing}).
    \item For the decided end-effector pose, a joint configuration of the manipulator is computed using an inverse kinematics solver with dynamic obstacle avoidance~\cite{kang2021rcik}.
\end{itemize}

The primary contributions of our work are highlighted as follows:
\begin{itemize}
    \item We present the first attempt to simultaneously address occlusion avoidance, collision avoidance, and object tracking for manipulator-driven visual tracking systems.
    \item Handling multiple objectives concurrently increases the risk of driving the manipulator to its motion range limits. The reachability measure is employed to ensure that the manipulator operates within a reachable range, which significantly reduces collisions and tracking failures.
\end{itemize}

To validate our proposed method, we conducted evaluations in various environments, including scenarios with a moving target object and varying velocities of dynamic obstacles. Through rigorous ablation studies based on object tracking and occlusion avoidance, we demonstrated that our approach, which considers obstacle avoidance and kinematic constraints, significantly improves the object tracking rate while reducing the occurrence of collisions compared to methods without these considerations.

The rest of this letter is organized as follows. The next section describes the problem statement, including our goals and challenges. Section~\ref{sec:method} introduces our method, followed by simulations and hardware experiments provided in Sections~\ref{sec:simulations} and~\ref{sec:hardware_experiments}, respectively. The conclusions are presented in Section~\ref{sec:conclusions}.


\section{Problem Definition}
\label{sec:3}

Our goal is to develop a camera navigation system utilizing a redundant manipulator, enabling effective tracking of a target object while avoiding collisions and occlusions caused by obstacles. The use of a redundant manipulator offers the advantage of multiple joint configurations for achieving a single end-effector pose~\cite{kang2022analysis}. This flexibility allows for continuous camera movement, thereby improving the inherent limitations of the camera, including its limited FoV (Fig.~\ref{fig:fov}) and possible occlusions by obstacles on its LoS (Fig.~\ref{fig:occlusion}).

In this letter, we mainly focus on enhancing manipulation stability by considering the manipulator's reachability. As the manipulator moves to track a target object, there is a possibility of collisions with moving individuals and surrounding objects (Fig.~\ref{fig:collision}). This factor significantly affects the performance of visual tracking. To tackle this challenge, we integrate an additional RGB-D camera to identify obstacles around the manipulator.

During the navigation to track a target object or avoid dynamic obstacles, the manipulator may be driven to a point that is unreachable (Fig.~\ref{fig:kinematic_constraint}) due to the manipulator's kinematic constraints, potentially leading to unsuccessful object tracking or collisions with dynamic obstacles. To mitigate the occurrence of such failures, we pay attention to the existence of multiple camera poses capable of tracking the target object. We actively move the manipulator toward the configurations with increased reachability.

In summary, our problem involves performing camera navigation to overcome the camera's limitations while considering manipulation stability (Fig.~\ref{fig:problems}). Addressing these aspects in real-time presents a non-trivial and complex challenge. To address this problem, we propose an optimization-based navigation approach to determine the desired end-effector pose, taking into account the aforementioned constraints.

To compute a joint configuration from the end-effector pose, we employ a real-time inverse kinematics (IK) solver that considers dynamic obstacle avoidance and continuity of joint configurations; specifically, we use the RCIK~\cite{kang2021rcik}. This solver takes on the burden of collision avoidance and joint space continuity, allowing our optimizer to focus on the desired end-effector pose.


\section{Method}
\label{sec:method}

Our method is comprised of several components. The next subsection describes the computational flow of these components, followed by an additional subsection providing the details of the optimization component.


\subsection{Computational Flow}
\label{sec:3_system_flow}

\begin{figure}[t]
	\vspace{0.15cm}
	\centering
	\includegraphics[width=3.25in]{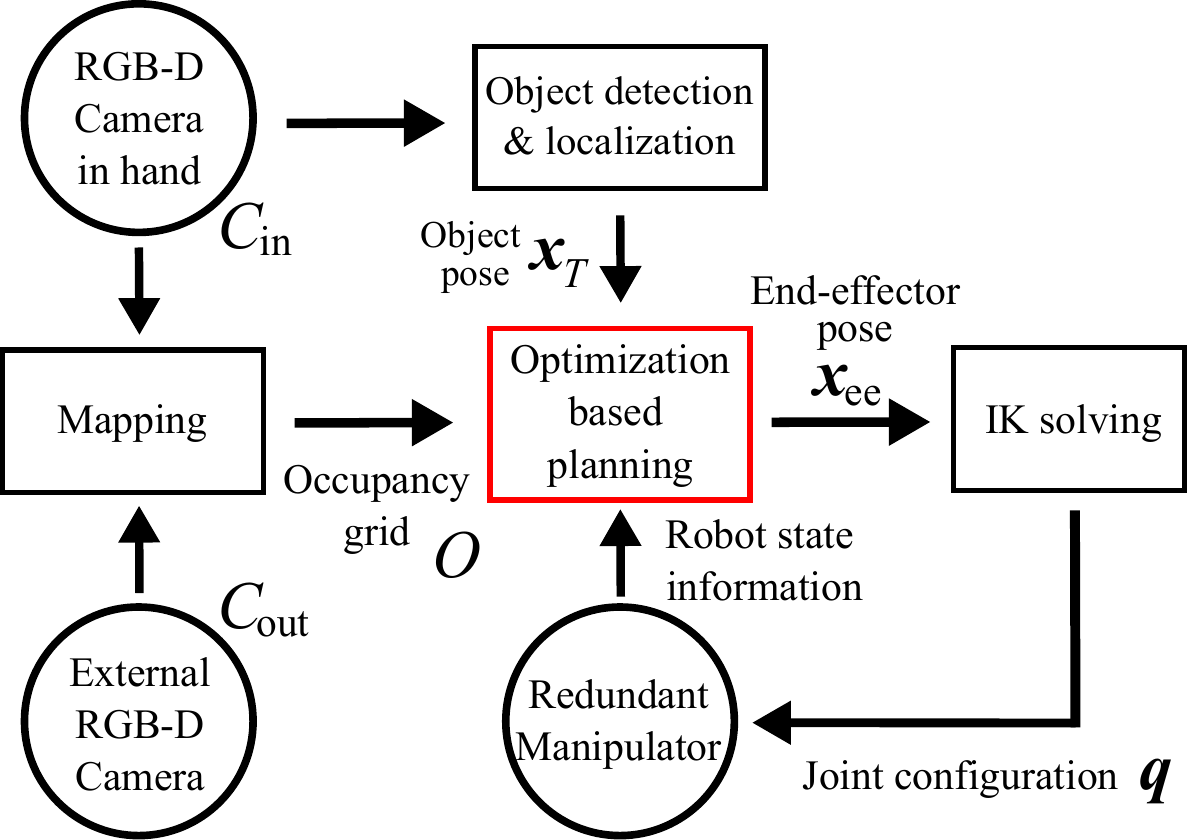}
	\caption{
		This figure shows our overall system flow.
	}
	\label{fig:system_flow}
	\vspace{-0.6cm}
\end{figure}

Fig.~\ref{fig:system_flow} illustrates the system flow of our optical tracking system. The system includes a redundant manipulator, an RGB-D camera, $C_\text{in}$, mounted on the manipulator's end-effector and another RGB-D camera, $C_\text{out}$, installed outside the manipulator to observe obstacles of potential collisions.
 
The target object, $T$, is detected and localized using $C_\text{in}$. Subsequently, the measurements of both cameras are combined to generate an occupancy grid, $O$, which is a binary voxel representation of space, distinguishing between vacant and occupied (or potentially occupied) spaces~\cite{kwon2019super} We exploit the occupancy grid to represent the obstacles as the occupied space. Our occupancy grid, $O$, provides two benefits: (i) reducing computational overhead and (ii) filtering out camera measurement noise. Additionally, we exclude the occupied cells corresponding to the manipulator and $T$, since these are not considered obstacles~\cite{kang2021rcik}.

We represent the pose of $T$ as a $6$D vector $\bm{x_{T}} \in \mathbb{R}^{6}$ by combining $3$D position and $3$D rotation vectors. The Euler XYZ angles were used for the rotation. Using the obstacle information in the occupancy grid, we determine the end-effector pose, $\bm{x_\text{ee}} \in \mathbb{R}^{6}$, using an optimization-based planner considering the four objectives in Fig.~\ref{fig:problems}. This component is discussed in detail in the following subsection. The determined $\bm{x_\text{ee}}$ is then used to control the manipulator by finding an appropriate joint configuration, $\bm{q}$, through the real-time IK solver~\cite{kang2021rcik}.


\subsection{Optimization-Based Planning}
\label{sec:4_objective_function}

\begin{figure*}[t]
	\vspace{0.3cm}
	\centering
	\subfigure [Object tracking] {
		\includegraphics[width=1.62in]{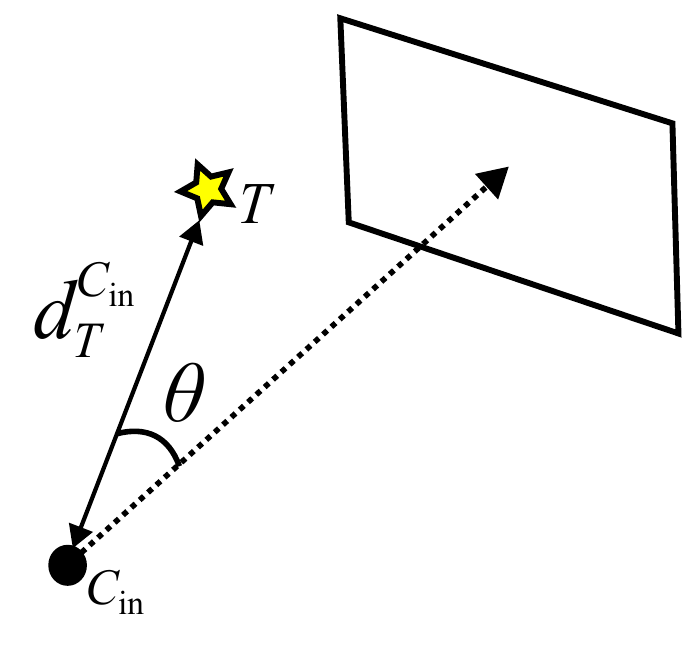}
		\label{fig:object_tracking}
	}
    \subfigure [Occlusion avoidance] {
		\includegraphics[width=1.62in]{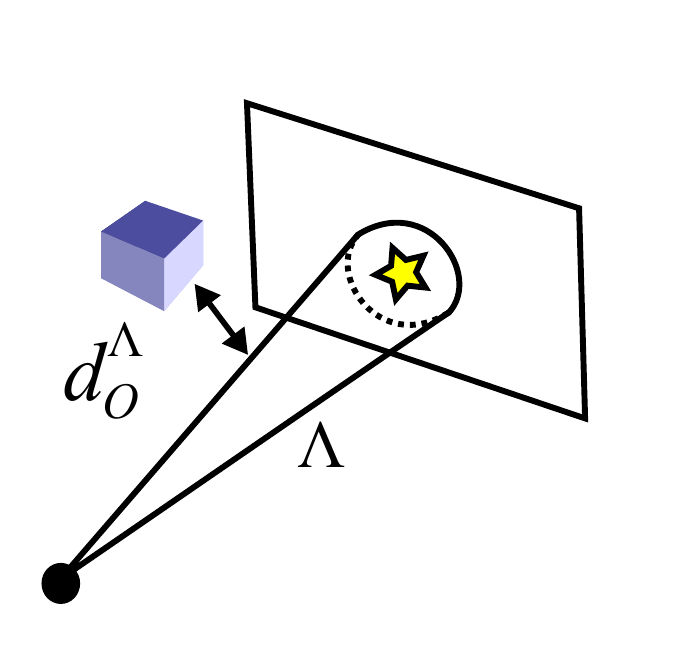}
		\label{fig:occlusion_avoidance} 
	} 
	\subfigure [Collision avoidance] { 
		\includegraphics[width=1.62in]{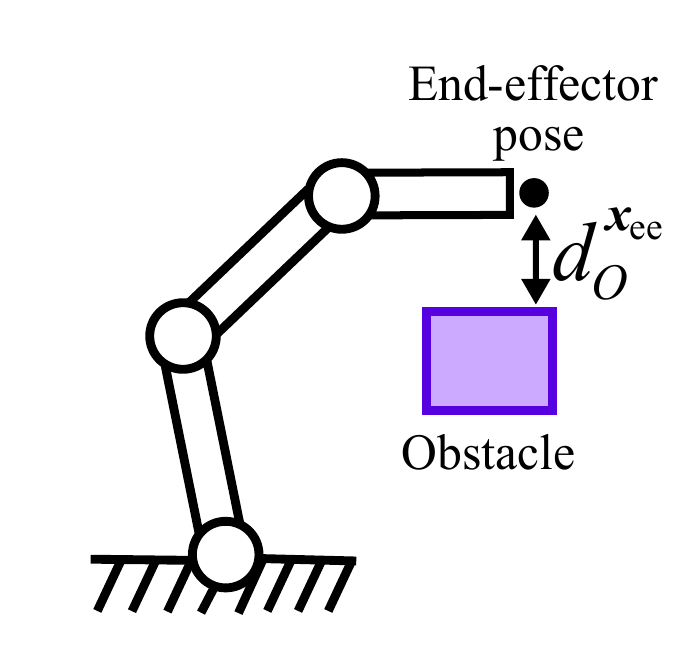}
		\label{fig:collision_avoidance} 
	}
	\subfigure [Reachability map] { 
		\includegraphics[width=1.62in]{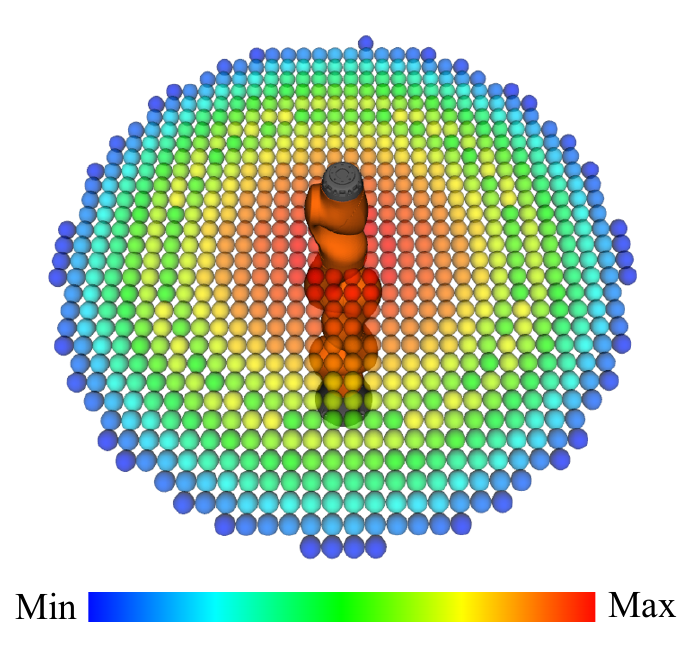}
		\label{fig:reachability} 
	}
	\caption{
            For objectives considered in our method.
	}
	\label{fig:objective}
	\vspace{-0.4cm}
\end{figure*}

We employ non-linear optimization to determine the desired end-effector pose $\bm{x_\text{ee}}$, as denoted by the red component in Fig.~\ref{fig:system_flow}.
During the optimization process, we consider four essential factors: object tracking, occlusion avoidance, collision avoidance, and the manipulator's reachability (see Fig.~\ref{fig:objective}). These factors are incorporated into our objective function, $\mathcal{U}$, as the sum of individual objectives:
\begin{equation}
    \mathcal{U} = F_\text{track} + F_\text{occl} + F_\text{col} + F_\text{reach}.
\end{equation}
Using $\mathcal{U}$, we optimize the change of end-effector pose, $\Delta \bm{x_\text{ee}}$, for a given time step $\Delta t$. Our optimization problem can be expressed as follows:
\begin{equation}
    \begin{split}
        \Delta \bm{x_\text{ee}} = \arg \min_{\Delta \bm{x_\text{ee}}} \mathcal{U}(\Delta \bm{x_\text{ee}}) \\
        \text{subject to } \bm{l} \leq \Delta \bm{x_\text{ee}} \leq \bm{u},
    \end{split}
    \label{eq:objective_function}
\end{equation}
where $\bm{l}$ and $\bm{u}$ are the lower and upper bounds of $\Delta \bm{x_\text{ee}}$.

Noting that each term in $\mathcal{U}$ rewards or penalizes certain system behaviors (e.g., penalizing for small distances to obstacles), a typical approach is to use a least-squares objective, where a quadratic function is used to rescale error terms. In our method, however, we use the following cubic function as a rescale function:
\begin{equation}
    G_{\mathbf{\bm{w}}} (x) = w_{0} (w_{1} x + w_{2})^3,
\end{equation}
where $x \in \Re$ is the system behavior to be rewarded or penalized, and $\mathbf{\bm{w}} = (w_0, w_1, w_2) \in \Re^3$ are user parameters.
This function is chosen for multiple reasons:
(i) Each term in $\mathcal{U}$ possesses distinct characteristics, which makes it challenging to control their relative importance. We empirically found in our previous work~\cite{kang2023ucarp} that this function performed nicely in handling relative scales.
(ii) There are cases where the sign of $x$ needs to be preserved (e.g., negative distance to represent penetration depth).
(iii) Tuning three coefficients $w_{0}$, $w_{1}$, and $w_{2}$ provides sufficient flexibility for user to design each term.
\newline

\subsubsection{Object tracking}
\label{sec:4_object_tracking}

To track $T$, it is essential to keep $T$ within the camera's FoV. Assuming that $T$ can move in arbitrary directions, it is desired to have $T$ at the center of the camera's FoV. Therefore, we move the manipulator to position $T$ at the center of the camera's FoV while maintaining a desired distance, $d_{\text{des}} \in \Re$, from the camera to the target.

In our method, we allow the camera to view $T$ from various angles because the specific angle at which the camera views the object is not of great importance. Advances in object locating and tracking technologies, as discussed in prior studies~\cite{ren2013marker, garrido2014automatic}, enable effective tracking from multiple angles. 

Overall, we define $F_\text{track}$ to penalize the distance and centering errors:
\begin{equation}
    F_\text{track}(\bm{x_\text{ee}}) = G_{\mathbf{\bm{w_{d}}}}(| d_{\text{des}} - d_{T}^{C_\text{in}} |) + G_{\mathbf{\bm{w_{\theta}}}}(\theta),
    \label{eq:obj_track}
\end{equation}
where $d_{T}^{C_\text{in}} \in \Re$ denotes the distance between $T$ and $C_\text{in}$, $d_{\text{des}} \in \Re$ denotes the desired $d_{T}^{C_\text{in}}$,
and $\theta \in [0, \pi]$ represents the angle of the target with respect to the camera's central view vector, as illustrated in Fig.~\ref{fig:object_tracking}.

\subsubsection{Occlusion avoidance}
\label{sec:4_occlusion_avoidance}

Obstacles obstructing the camera's view give a significant challenge to object tracking. To address this issue, various methods~\cite{wang2015robot, meng2021development} represented the camera's sight for a target object as a cone, $\Lambda$, and made efforts to ensure $\Lambda$ is clear without obstacles. In our method, we calculate the minimum distance between $\Lambda$ and the occupancy grid $O$ and use it to adjust the camera's pose to ensure that the target object remains unobstructed. Accordingly, $F_\text{occl}$ is defined as follows:
\begin{equation}
    F_\text{occl}(\bm{x_\text{ee}}) = 
    \begin{cases}
        G_{\mathbf{\bm{w_{\text{occl}}}}}(d_{O}^{\Lambda}), & \text{if $ d_{O}^{\Lambda} < u_{\text{occl}}$},\\
        0, & \text{otherwise},
    \end{cases}
    \label{eq:obj_occl}
\end{equation}
where $d_{O}^{\Lambda} \in \Re$ represents the minimum distance between $O$ and $\Lambda$ (see Fig.~\ref{fig:occlusion_avoidance}), and $u_{\text{occl}} \in \Re$ is a deactivation distance threshold. It is worth noting that $d_{O}^{\Lambda}$ is a signed distance value, where the negative distance is computed as the penetration depth when the occupancy grid interferes with $\Lambda$. When obstacles are sufficiently far (i.e., when $ d_{O}^{\Lambda} < u_{\text{occl}}$), occlusion avoidance is deactivated by setting $F_\text{occl}(\bm{x_\text{ee}}) = 0$.

\subsubsection{Collision avoidance}
\label{sec:4_collision_avoidance}

In the optimization process, we only concentrate on the end-effector's obstacle avoidance. We remark that the collision avoidance of the manipulator body is taken care of in the IK solver, of which the details can be found in \cite{rakita2021collisionik, kang2021rcik}.

In alignment with several methods~\cite{hauser2012responsiveness, rakita2021collisionik, kang2021rcik} considering dynamic obstacle avoidance, we utilize the distance to obstacles to guide the end-effector away from potential collisions.
Consequently, $F_\text{col}$ is defined using the minimum distance between the end-effector pose $\bm{x_\text{ee}}$ and obstacles in the occupancy grid $O$:
\begin{equation}
    F_\text{col}(\bm{x_\text{ee}}) =
    \begin{cases}
        G_{\mathbf{\bm{w_\text{col}}}}(d_{O}^{\bm{x_\text{ee}}}), & \text{if $ d_{O}^{\bm{x_\text{ee}}} < u_{\text{col}}$},\\
        0, & \text{otherwise},
    \end{cases}
    \label{eq:obj_col}
\end{equation}
where $d_{O}^{\bm{x_\text{ee}}}$ represents the minimum distance between $O$ and $\bm{x_\text{ee}}$ as illustrated in Fig.~\ref{fig:collision_avoidance}, and $u_{\text{col}} \in \Re$ is the distance threshold to deactivate collision avoidance.

\subsubsection{Reachability}
\label{sec:4_reachability}

During the process of avoiding collision and occlusion and tracking a moving object simultaneously, the manipulator could encounter its kinematic limits, potentially leading to failures in collision avoidance or object tracking. To address this issue, we utilize a reachability map~\cite{zacharias2007capturing}, which consists of a pre-computed collection of reachable end-effector poses arranged in a $3$D grid structure. Grids with denser end-effector poses possess higher reachability scores.
In practice, the reachability map serves as an indicator of how easily the specific end-effector pose can be reached.
It has been employed to determine if a target end-effector pose is reachable or to find a suitable base pose for reaching the target~\cite{zacharias2007capturing, makhal2018reuleaux, kang2019harmonious}.

We incorporate the reachability map into our optimization process to prevent the manipulator from approaching unreachable regions.
The reachability map is constructed offline by evaluating the number of available IK solutions for each grid cell, considering various uniformly sampled orientations. Consequently, the value assigned to a cell reflects the ratio of successful IK solutions to the number of attempts, with self-collision checks included. As an example, a layer of the reachability map is visualized in Fig.~\ref{fig:reachability}. Using the reachability map, we define the function $F_\text{reach}$ to penalize end-effector poses with low reachability:
\begin{equation}
    F_\text{reach}(\bm{x_\text{ee}}) =
    \begin{cases}
        G_{\mathbf{\bm{w_\text{reach}}}}(V(\bm{x_\text{ee}})), & \text{if $V(\bm{x_\text{ee}}) < u_{\text{reach}}$},\\
        0, & \text{otherwise},
    \end{cases}
    \label{eq:obj_track}
\end{equation}
where $V(\bm{x_\text{ee}}) \in \Re$ is the reachability score of $\bm{x_\text{ee}}$, and $u_{\text{reach}} \in \Re$ is the deactivation reachability threshold. The discrete grid values of $V(\bm{x_\text{ee}})$ are interpolated to provide continuous function reachability scores.


\section{Simulations}
\label{sec:simulations}

In this section, we discuss our simulation results with a description of our experimental setting.
We performed simulations on a machine with a $3.60$ GHz Intel i$9$-$9900$K CPU.

\subsection{Simulation Environment}

\begin{figure}[t]
	\vspace{0.5cm}
	\centering 
	\subfigure [Scenario 1] {
		\includegraphics[width=1.61in]{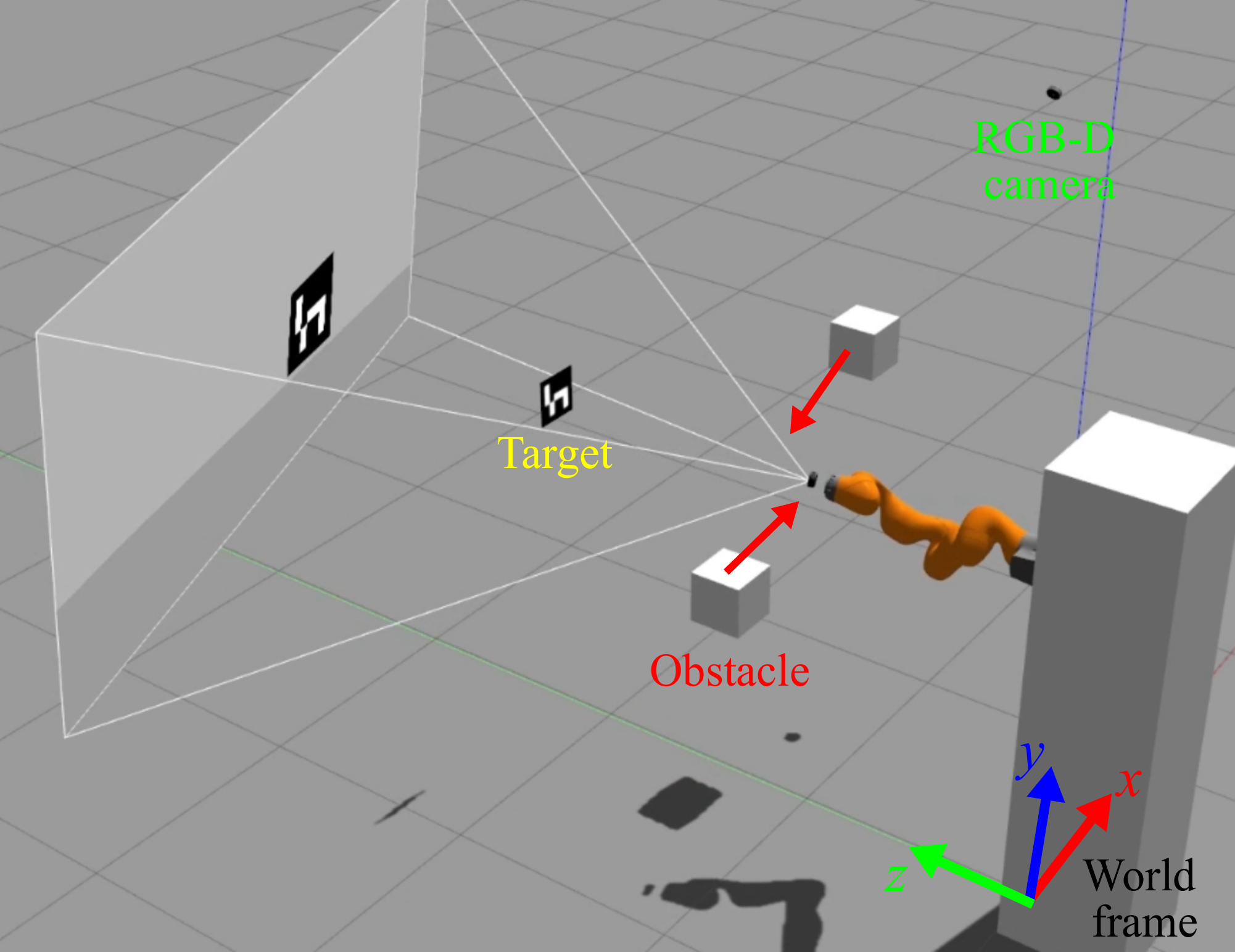}
		\label{fig:scene_1} 
	} 
	\subfigure [Scenario 2] { 
		\includegraphics[width=1.54in]{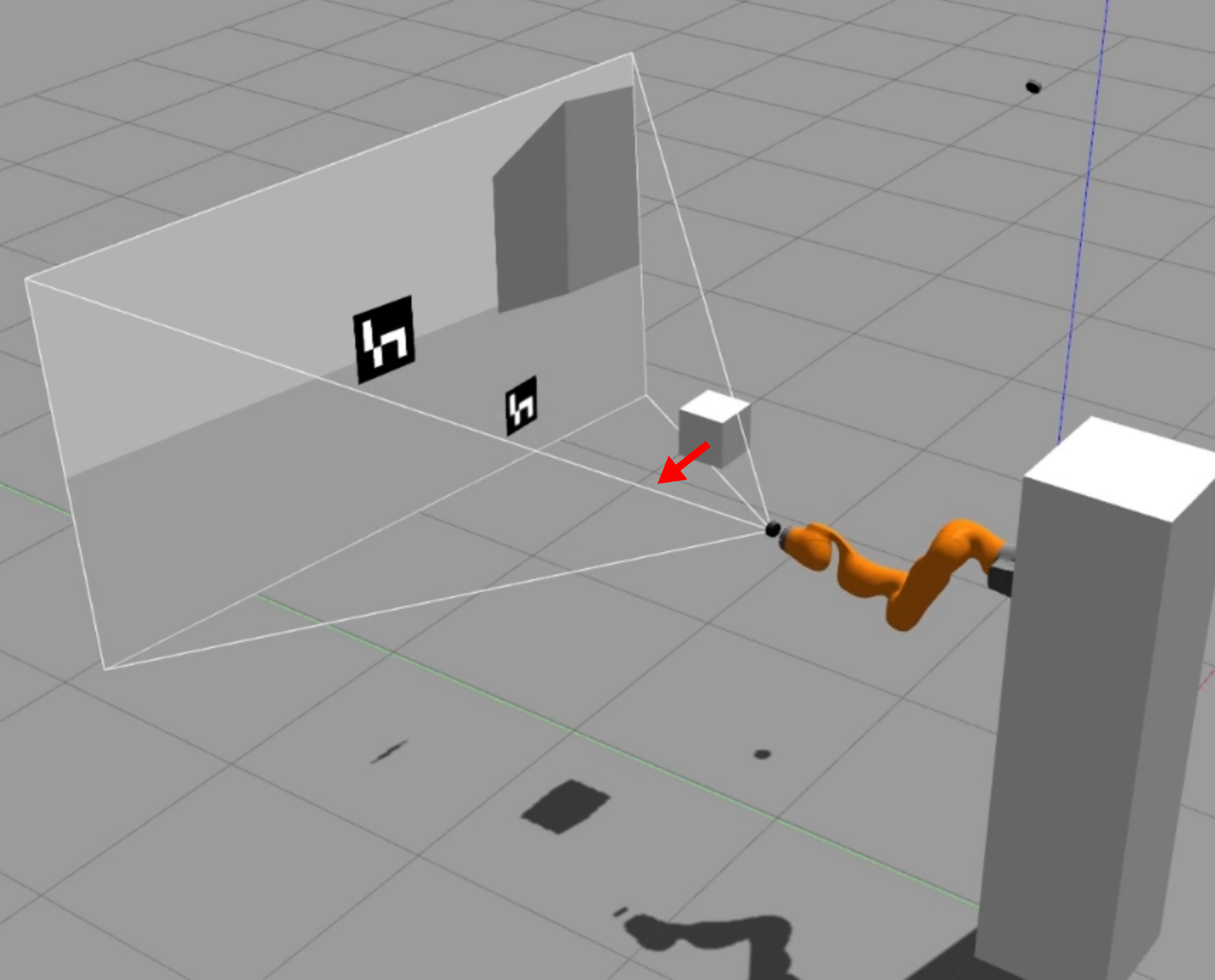}
		\label{fig:scene_2} 
	}
	\caption{
            Two scenarios considered in our simulations. In scenario 1 depicted in (a), obstacles continuously passed between the manipulator and the target object, potentially leading to collisions. In scenario 2 in (b), an obstacle stopped after moving, obstructing the camera's view of the target.
	} 
	\label{fig:test_scenes}
	\vspace{-0.6cm}
\end{figure}

A ROS environment was developed, where a $7$-DoF KUKA iiwa manipulator was used with an RGB-D camera (RealSense L515) mounted on its end-effector (see Fig~\ref{fig:test_scenes}).  Additionally, we incorporated an external RGB-D camera to detect approaching obstacles. 

Using the measurements of the two RGB-D cameras, we generated an occupancy grid using the SuperRay method~\cite{kwon2019super}. The occupancy grid was updated within $50$ ms on average. In parallel, we employed the ArUco marker to detect and localize the target objects, the process of which took approximately $2$ ms; when the target is missed due to occlusions or tracking failures, we use the most recent target position. Lastly, we allocated $30$ ms to the IK solver (RCIK~\cite{kang2021rcik}), securing approximately $20$ ms for our optimization process. We used sequential quadratic programming~\cite{kraft1988software} as a non-linear optimization solver. Further details regarding the parameter settings for the experiments can be found in Table~\ref{tab:parameters}.

\begin{table}[t]
	\vspace{0.5cm}
	\centering
	\scriptsize
	\renewcommand \arraystretch{1.4}
	\setlength{\tabcolsep}{1.8pt}
	\caption{
		Parameters in our experiment.
	}
	\begin{center}
		\begin{tabular}{|c|c|} 
			\hline
			\textbf{~Parameters~} & \textbf{Values} \\
			\hline
                $\bm{l}$ & ~~~[$-0.05$ m, $-0.05$ m, $-0.05$ m, $-0.2$ rad, $-0.2$ rad, $-0.2$ rad]~~~ \\
                $\bm{u}$ & [$0.05$ m, $0.05$ m, $0.05$ m, $0.2$ rad, $0.2$ rad, $0.2$ rad] \\
			\hline
                $\mathbf{\bm{w_{d}}}$, $\mathbf{\bm{w_{\theta}}}$, $d_{\text{des}}$ & $[0.5,\ 1.0,\ 0]$, $[7.5,\ 1.5,\ 0]$, $1.0$ m \\
			\hline
                $\mathbf{\bm{w_{\text{occl}}}}$, $u_{\text{occl}}$ & $[-1.0,\ 5.0,\ -1.5]$, $0.3$ m \\
			\hline
                $\mathbf{\bm{w_{col}}}$, $u_{\text{col}}$ & $[-1.0,\ 1.5,\ -1.5]$, $1.0$ m \\
			\hline
                ~~$\mathbf{\bm{w_{reach}}}$, $u_{\text{reach}}$~~ & $[-5.0, \ 100.0, \ -50.0]$, $0.5$ \\
			\hline
		\end{tabular}
	\end{center}
	\vspace{-0.4cm}
	\label{tab:parameters}
\end{table}

\subsection{Scenarios}
\label{sec:5_experimental_setting}

To assess the effectiveness of our method, we conducted experiments in two different scenarios:

\noindent {\bf Scenario 1:}
The first scenario is depicted in Fig~\ref{fig:scene_1}, where we included varying numbers of obstacles, ranging from $0$ to $2$, that passed between the manipulator and the target. The target moved randomly within a boxed boundary, from $[-1.0 \text{ m}, 1.8 \text{ m}, 1.0 \text{ m}]$ to $[-1.0 \text{ m}, 2.4 \text{ m}, 2.0 \text{ m}]$ relative to the world frame as shown in Fig~\ref{fig:scene_1}. In this scenario, we examined whether our method could successfully track the target while avoiding obstacles. 

\noindent  {\bf Scenario 2:}
In the second scenario as shown in Fig~\ref{fig:scene_2}, we prepared a different challenge where an obstacle obstructed the camera's view of the target, and the target was either in motion or stationary. In this case, we evaluated how effectively our method handled occlusions.

Overall, we tested five distinct cases, three arising in the first scenario by the varying number of obstacles and two in the second scenario by the motion of the target. To assess each case, we performed $50$ tests for each of them. In each test, we followed a sequence of $100$ time steps, with the time step chosen to be $50$ ms for occupancy grid construction.

In our simulations, we set a constant velocity for the target as $0.5$ m$/$s. The obstacles had varying velocities within the range of $0.8$ m$/$s to $1.2$ m$/$s. Since the median velocity of the obstacle is equal to the maximum velocity of the robot's end-effector in each spatial axis (i.e., $1.0$ m$/$s based on $\bm{l}$ and $\bm{u}$ in Table~\ref{tab:parameters}), we could see how effectively our method avoided the obstacle's obstruction while tracking the target.

\begin{figure*}[t]
	\vspace{0.3cm}
	\centering
	\subfigure [Collision avoidance] {
		\includegraphics[width=6.9in]{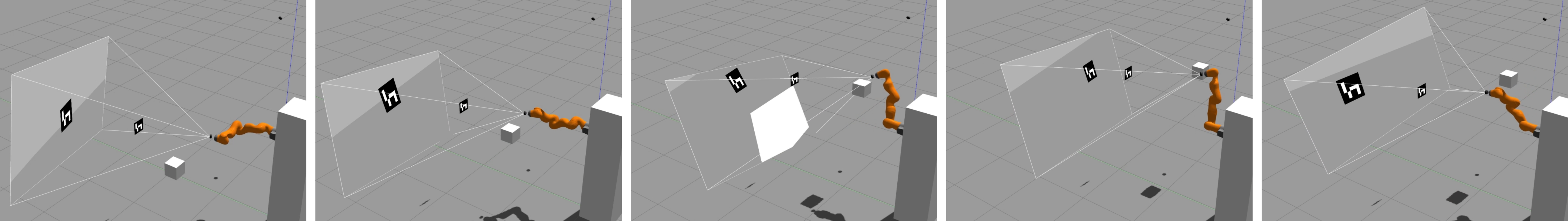}
		\label{fig:res_col} 
	} 
	\subfigure [Occlusion avoidance] { 
		\includegraphics[width=6.9in]{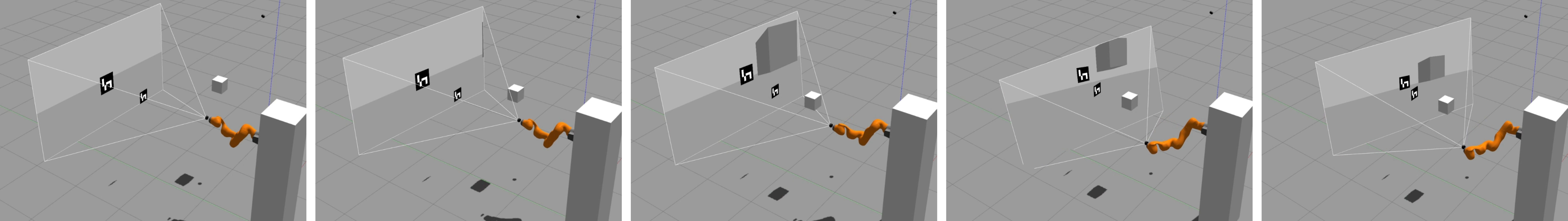}
		\label{fig:res_occl} 
	}
	\caption{
            Example motions generated by our method demonstrating (a) collision avoidance and (b) occlusion avoidance. In both situations, the manipulator tracked the target object while avoiding obstacles and occlussions. See our video attachment for more details.
	}
	\label{fig:res_seq}
	\vspace{-0.4cm}
\end{figure*}

\begin{table}[t]
    \vspace{0.5cm}
    \centering
    \scriptsize
    \renewcommand \arraystretch{1.4}
    \setlength{\tabcolsep}{1.8pt}
    \caption{
        Results for various configurations of $\mathcal{U}$ and different numbers of obstacles in scenario 1.
    }
    \begin{center}
        \begin{tabular}{|c|c|c|c|c|c|} 
            \hline
            \multirow{2}{*}{\# of} & ~ & \# of & Average \# & \multirow{2}{*}{IK failure} & \multirow{2}{*}{Tracking} \\
            \multirow{2}{*}{obstacles} & $\mathcal{U}$ & failures by & of elapsed & \multirow{2}{*}{rate} & \multirow{2}{*}{rate} \\
            ~ &  & collisions & time steps & ~ & ~ \\
            \hline
            \hline
            ~~ & $F_\text{track} + F_\text{occl}$ & 0 & 100 & 0.44 & 0.82 \\
            \cline{2-6}
            ~\multirow{2}{*}{0}~ & $F_\text{track} + F_\text{occl} + F_\text{col}$ & 0 & 100 & 0.41 & 0.83 \\
            \cline{2-6}
            ~~ & $F_\text{track} + F_\text{occl} + F_\text{reach}$ & 0 & 100 & \textbf{0.04} & \textbf{0.98} \\
            \cline{2-6}
            ~ & $F_\text{track} + F_\text{occl} + F_\text{col} + F_\text{reach}$ & 0 & 100 & 0.06 & 0.97 \\
            \hline
            \hline
            ~~ & $F_\text{track} + F_\text{occl}$ & 41 & 40.6 & 0.32 & 0.90 \\
            \cline{2-6}
            ~\multirow{2}{*}{1}~ & $F_\text{track} + F_\text{occl} + F_\text{col}$ & 33 & 53.7 & 0.67 & 0.88 \\
            \cline{2-6}
            ~~ & $F_\text{track} + F_\text{occl} + F_\text{reach}$ & 43 & 40.2 & \textbf{0.05} & \textbf{0.95} \\
            \cline{2-6}
            ~ & $F_\text{track} + F_\text{occl} + F_\text{col} + F_\text{reach}$ & \textbf{9} & \textbf{90.2} & 0.17 & 0.93 \\
            \hline
            \hline
            ~~ & $F_\text{track} + F_\text{occl}$ & 47 & 39.6 & 0.36 & 0.91 \\
            \cline{2-6}
            ~\multirow{2}{*}{2}~ & $F_\text{track} + F_\text{occl} + F_\text{col}$ & 30 & 62.4 & 0.64 & 0.88 \\
            \cline{2-6}
            ~~ & $F_\text{track} + F_\text{occl} + F_\text{reach}$ & 45 & 44.2 & \textbf{0.04} & \textbf{0.93} \\
            \cline{2-6}
            & $F_\text{track} + F_\text{occl} + F_\text{col} + F_\text{reach}$ & \textbf{12} & \textbf{92.1} & 0.16 & 0.91 \\
            \hline
        \end{tabular}
    \end{center}
    \vspace{-0.4cm}
    \label{tab:results_scenario_1}
\end{table}

\begin{table}[t]
    \vspace{0.5cm}
    \centering
    \scriptsize
    \renewcommand \arraystretch{1.4}
    \setlength{\tabcolsep}{1.8pt}
    \caption{
        Results for various configurations of $\mathcal{U}$ in scenario 2, where the camera view of either a stationary or moving target was obstructed by an obstacle.
    }
    \begin{center}
        \begin{tabular}{|c|c|c|c|c|c|} 
            \hline
            ~ & ~ & \# of & Average \# & \multirow{2}{*}{IK failure} & \multirow{2}{*}{Tracking} \\
            ~$T$~ & $\mathcal{U}$ & failures by & of elapsed & \multirow{2}{*}{rate} & \multirow{2}{*}{rate} \\
             &  & collisions & time steps & ~ & ~ \\
            \hline
            \hline
            ~~ & $F_\text{track} + F_\text{occl}$ & 3 & 95.4 & 0.74 & 0.50 \\
            \cline{2-6}
            ~\multirow{2}{*}{Stop}~ & $F_\text{track} + F_\text{occl} + F_\text{col}$ & \textbf{0} & \textbf{100} & 0.86 & 0.77 \\
            \cline{2-6}
            ~~ & $F_\text{track} + F_\text{occl} + F_\text{reach}$ & \textbf{0} & \textbf{100} & 0.10 & 0.77 \\
            \cline{2-6}
            ~ & $F_\text{track} + F_\text{occl} + F_\text{col} + F_\text{reach}$ & \textbf{0} & \textbf{100} & \textbf{0.06} & \textbf{0.90} \\
            \hline
            \hline
            ~~ & $F_\text{track} + F_\text{occl}$ & 6 & 90.9 & 0.55 & 0.71 \\
            \cline{2-6}
            ~\multirow{2}{*}{Move}~ & $F_\text{track} + F_\text{occl} + F_\text{col}$ & 1 & 98.7 & 0.72 & 0.69 \\
            \cline{2-6}
            ~~ & $F_\text{track} + F_\text{occl} + F_\text{reach}$ & \textbf{0} & \textbf{100} & \textbf{0.11} & 0.89 \\
            \cline{2-6}
            ~ & $F_\text{track} + F_\text{occl} + F_\text{col} + F_\text{reach}$ & \textbf{0} & \textbf{100} & 0.25 & \textbf{0.90} \\
            \hline
        \end{tabular}
    \end{center}
    \vspace{-0.4cm}
    \label{tab:results_scenario_2}
\end{table}

\begin{figure}[t]
	\vspace{0.1cm}
	\centering 
	\subfigure [With one obstacle] {
		\includegraphics[width=1.58in]{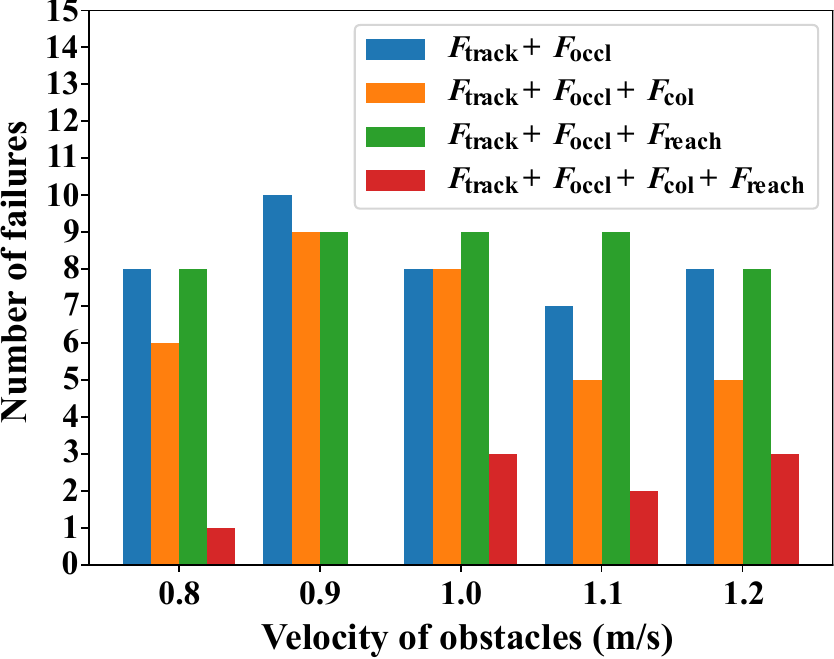}
		\label{fig:plot_vel_1} 
	} 
	\subfigure [With two obstacles] { 
		\includegraphics[width=1.58in]{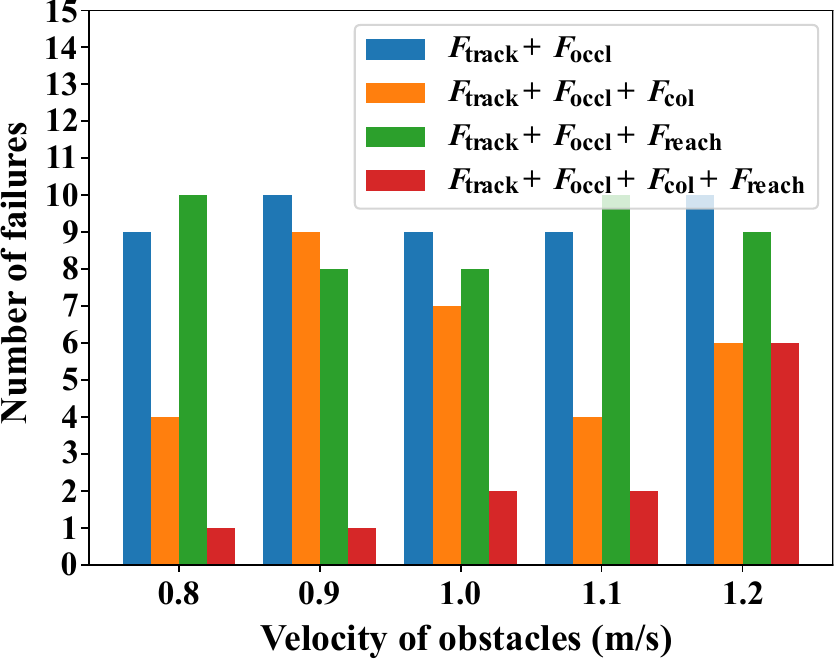}
		\label{fig:plot_vel_2} 
	}
	\caption{
            Histograms of failure counts by collisions with respect to velocities of obstacles in the presence of (a) one obstacle and (b) two obstacles.
	} 
	\label{fig:plot_vel}
	\vspace{-0.6cm}
\end{figure}

\subsection{Evaluation Criteria}
\label{sec:5_evaluation}

We analyzed the collision avoidance performance by measuring two critical performance metrics: (i) the number of failures by collisions and (ii) the average number of elapsed time steps. A failure was recorded when the manipulator collided with obstacles, and in such instances, the number of time steps only elapsed until the collision occurred. These metrics provided valuable insights into how effectively the tested method avoids obstacles.

Additionally, we assessed two other key metrics regarding manipulation stability and tracking performance: (iii) the average IK failure rate and (iv) the average tracking rate over the elapsed time steps. The IK failure rate indicated how often the tested method failed to generate an end-effector pose that satisfied the manipulator's kinematic constraints. On the other hand, the tracking rate measured the rate of successful detection of the target. Note that both metrics were evaluated over the elapsed time steps, not the total of $100$ time steps.

\subsection{Results}

Table~\ref{tab:results_scenario_1} and~\ref{tab:results_scenario_2} show the overall results of our experiments. Example motions are visually provided in Fig.~\ref{fig:res_seq}. In Fig.~\ref{fig:res_col}, the manipulator avoided an approaching obstacle, while keeping the camera view toward the target. In Fig.~\ref{fig:res_col}, the manipulator steered the camera so that a potential occlusion was avoided, as particularly observable in the third and fourth snapshots. We recommend watching our video attachment for more detailed motions.

We conducted an ablation study on our method. Since the prior methods~\cite{wang2015robot, meng2021development} considered object tracking and occlusion avoidance, the ablation study was based on $\mathcal{U} = F_\text{track} + F_\text{occl}$. Next, we tested our approach with additional considerations for collision avoidance ($F_\text{col}$) against external obstacles and reachability ($F_\text{reach}$). In this section, we refer to $\mathcal{U} = F_\text{track} + F_\text{occl} + F_\text{col} + F_\text{reach}$ as our method and $\mathcal{U} = F_\text{track} + F_\text{occl}$ as the baseline approach.

In Table~\ref{tab:results_scenario_1} and~\ref{tab:results_scenario_2}, we can observe the impact of including $F_\text{col}$ and $F_\text{reach}$. Incorporating both terms together led to a reduction in the number of failures by collisions, resulting in an increased number of elapsed time steps. On the other hand, when $F_\text{col}$ was added alone to the baseline approach, it showed the highest IK failure rate and the lowest tracking rate in the presence of obstacles. Moreover, there were only slight improvements compared to the baseline approach in terms of the number of failures by collisions and elapsed time steps. This outcome can be attributed to the manipulator being driven to an inaccessible area while moving away from obstacles; note that the manipulator did not move until finding a solution.

\begin{figure*}[t!]
	\vspace{0.3cm}
	\centering
	\subfigure [Moving obstacle ($\mathcal{U} = F_\text{track} + F_\text{occl} + F_\text{col}$)] {
    		\includegraphics[width=6.9in]{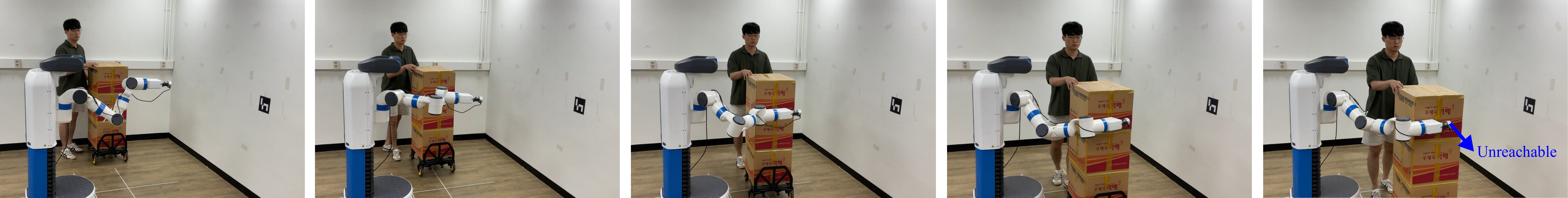}
    		\label{fig:res_real_obs_wo_reach} 
	} 
	\subfigure [Moving obstacle ($\mathcal{U} = F_\text{track} + F_\text{occl} + F_\text{col} + F_\text{reach}$)] { 
    		\includegraphics[width=6.9in]{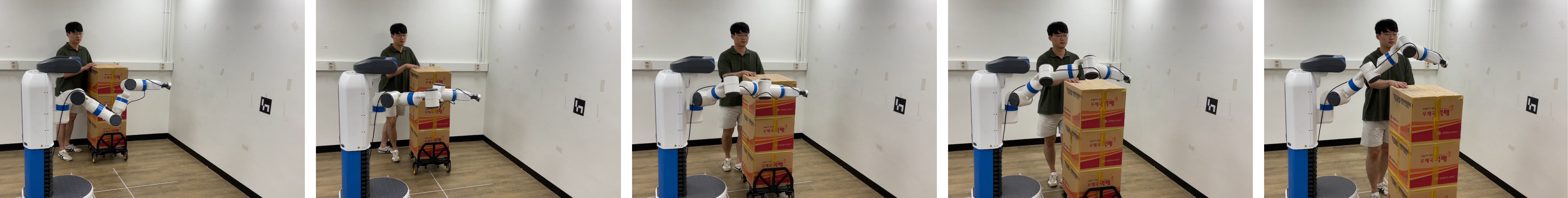}
    		\label{fig:res_real_obs_w_reach} 
	}
        \subfigure [Moving target object ($\mathcal{U} = F_\text{track} + F_\text{occl} + F_\text{col}$)] { 
    		\includegraphics[width=6.9in]{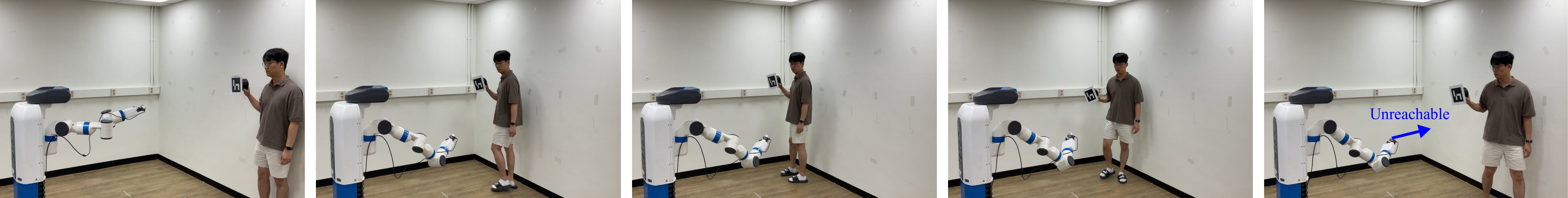}
    		\label{fig:res_real_tar_wo_reach} 
	}
        \subfigure [Moving target object ($\mathcal{U} = F_\text{track} + F_\text{occl} + F_\text{col} + F_\text{reach}$)] { 
    		\includegraphics[width=6.9in]{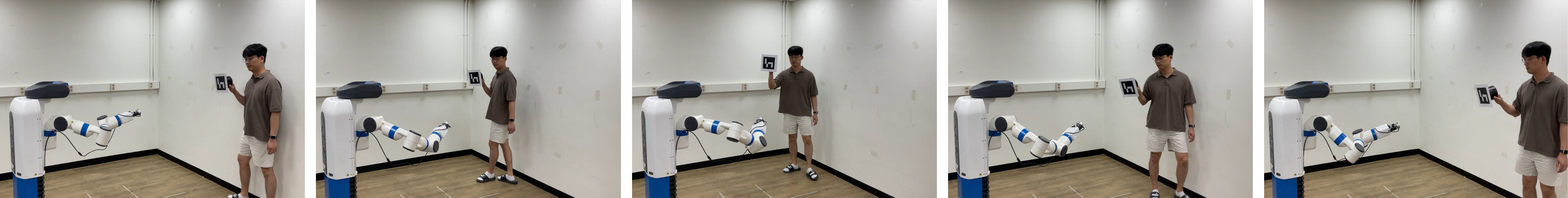}
    		\label{fig:res_real_tar_w_reach} 
	}
	\caption{
            Experimental motions generated using various configurations of $\mathcal{U}$ for stationary and moving targets. The motions in (a) and (c) were generated without including $F_\text{reach}$, which resulted in a collision and a tracking failure due to kinematic limits, respectively. The motions in (b) and (d) were generated by incorporating $F_\text{reach}$ and did not experience such issues. See our video attachment for detailed motions.
	}
	\label{fig:res_real_seq}
	\vspace{-0.4cm}
\end{figure*}

The inclusion of $F_\text{reach}$ yielded a notable reduction in the IK failure rate, as it effectively prevented the manipulator from reaching inaccessible areas. This result confirms that $F_\text{reach}$ aids the optimizer in generating an end-effector pose that satisfies the manipulator's kinematic constraints. The utilization of the reachability map successfully guided the manipulator's movement toward areas with high probabilities of IK success, ultimately leading to the observed decrease in the IK failure rate.
Consequently, our method demonstrated remarkable performance in handling overall challenges, with the lowest number of failures and the highest number of elapsed time steps, while maintaining the tracking rate sufficiently high.

We additionally analyzed the results with respect to the velocities of obstacles. As depicted in the histograms in Fig.~\ref{fig:plot_vel}, the number of failures by collisions increased as the obstacle velocity increased.
Note that it is essential to acknowledge that achieving a perfect solution for avoiding dynamic obstacles is challenging~\cite{vannoy2008real, lehner2015incremental}. Additionally, as the velocities of dynamic obstacles increase, the collision avoidance task becomes exponentially difficult. Nevertheless, our method exhibited significantly smaller numbers of failures compared to the other cases with partial objective functions.

In Table~\ref{tab:results_scenario_2}, our method demonstrated superior tracking performance compared to the baseline approach, even in scenarios with occlusion. While the baseline approach considered occlusion avoidance, its tracking rate was notably low due to frequent IK failures.
Furthermore, the baseline approach experienced failures a few times due to collisions with external obstacles, even though we intended the obstacle to pass between the target and the manipulator at its initial configuration.

In summary, our method has been validated through a series of simulations, showcasing its ability to enhance the stability of the manipulator by incorporating external obstacle avoidance. Additionally, the method has proven to be effective in accurately tracking the target while taking into account the kinematic constraints of the manipulator. These results support the practicality and reliability of our approach for improving both stability and tracking performance.


\section{Hardware Experiments}
\label{sec:hardware_experiments}

We performed a set of hardware experiments with the real Fetch manipulator to validate the practicality of our approach in real-world scenarios. We set up two distinct cases for evaluation: (i) One was that the robot tracked a fixed target object in a situation where a dynamic obstacle approached. (ii) The other was that the robot continuously tracked a moving target object. In addition, we assessed the importance of incorporating $F_\text{reach}$ by comparing results when $F_\text{reach}$ was and was not included in $\mathcal{U}$.

The comparisons are presented in Fig~\ref{fig:res_real_seq}. Once again, we refer to our video attachment for more detailed motions. As shown in Fig~\ref{fig:res_real_obs_w_reach}, our method successfully tracked a target object while avoiding obstacles. Conversely, in Fig~\ref{fig:res_real_obs_wo_reach} where $F_\text{reach}$ was not considered, a collision occurred as it went out of the reachable range of the manipulator.
These findings held true even in an obstacle-free environment. When tracking a moving target object, there were instances where the target traveled in a direction that the manipulator could not steer the camera toward, as shown in Fig~\ref{fig:res_real_tar_wo_reach}.

These results highlight the significance of considering kinematic constraints together with collision avoidance when generating manipulator motions for visual tracking tasks.


\section{Conclusions}
\label{sec:conclusions}

In this letter, we presented an optimization-based navigation method designed for a manipulator-driven visual tracking system in environments including dynamic obstacles. Our method addressed key challenges, including object tracking, occlusion avoidance, collision avoidance, and kinematic constraints. Through comprehensive evaluations, we have demonstrated the effectiveness of our method through simulations and hardware experiments.

Our simulation and experimental results showed that the consideration of collision avoidance solely in a nominal visual tracking approach could lead to increased IK and tracking failures and potential collisions, primarily because the manipulator might be driven to its kinematic limits. Our method addressed this issue by incorporating a reachability measure, which greatly enhanced manipulation stability with significantly improved success rates in collision avoidance, IK, and visual tracking.


    \section*{Acknowledgment}

    The authors would like to thank Prof. Sung-Eui Yoon and the members of SGVR Lab at KAIST for granting us permission to use their Fetch manipulator in our experiments.
        
    {
        \small
        \bibliographystyle{ieee/ieee}
        \bibliography{./mc}
    }

\end{document}